\documentclass[sigconf]{acmart}

\usepackage{booktabs}
\graphicspath{{figures/}}

\setcopyright{none}
\renewcommand\footnotetextcopyrightpermission[1]{}
\acmConference[AI4F '26]{3rd Workshop on LLMs and Generative AI for Finance}
{November 14, 2026}{Milan, Italy}

\acmYear{2026}

\newcommand{\CorpusIntro}{The corpus comes from NewsWitch, a
commercial financial-news product developed by Zanista AI Ltd that has run in daily production
since the beginning of 2024.\footnote{For access to NewsWitch data,
contact info@zanista.ai.}}
\newcommand{\CorpusPrior}{ The corpus has supported earlier studies
of LLM-driven trading and retrieval
\cite{kargarzadeh2024trading,ghatak2025alpha,khaledian2025pcarag}.}

\AtBeginDocument{\raggedbottom}

\begin{document}

\title{Buy the Rumor, Sell the News: When Is News Priced In?}

\author{Alireza Kargarzadeh}
\authornote{Corresponding author.}
\affiliation{%
  \institution{TailState Intelligence Ltd}
  \city{London}
  \country{United Kingdom}}
\email{alireza.kargarzadeh@tailstate.ai}

\author{Nariman Khaledian}
\affiliation{%
  \institution{Independent Researcher}
  \city{Antibes}
  \country{France}}
\email{nariman.khaledian@gmail.com}

\author{Navid Parvini}
\affiliation{%
  \institution{Zanista AI Ltd}
  \city{London}
  \country{United Kingdom}}
\email{navid.parvini@zanista.ai}

\author{Sid Ghatak}
\affiliation{%
  \institution{Increase Alpha, LLC}
  \city{Miami}
  \country{United States}}
\email{s.ghatak@increasealpha.com}

\author{Arman Khaledian}
\affiliation{%
  \institution{Zanista AI Ltd}
  \city{London}
  \country{United Kingdom}}
\email{arman.khaledian@zanista.ai}

\begin{abstract}
Two old market sayings hold that news is already priced in by the time it is published, and that the rumor is bought while the news is sold. Both are empirical claims about event time: they place the price move associated with a piece of news before and at publication rather than after it. Whether the claims hold, for which kinds of news, and by how much are basic questions about how fast markets absorb public information. We test them on 4.57 million financial news articles covering roughly 3{,}000 US stocks (2023--2026). A large language model teacher, distilled into a compact classifier through active learning, assigns each article one of 17 event tags and five attributes; articles are clustered into stories to separate first reports from follow-up coverage; and beta-adjusted abnormal returns are measured around the resulting 1.68 million stock-day events, with 364{,}405 neutral-sentiment events as a placebo group. Three results follow. First, the price move associated with news concentrates before and at publication: pooled across all signed events, the cumulative move in the news direction by the close of publication day is 2.8 times its value twenty days later, and for rumor-flagged events the directional return is concentrated around the rumor event, with little incremental return at the subsequent linked non-rumor event. Second, measured against the placebo of comparable stocks, markets underreact to numbers and overreact to stories: quantified fundamental news (earnings, dividends, guidance, analyst actions) keeps drifting in the direction of the news for weeks, while soft story-driven news (launches, macro commentary, leadership) gives back its move. Third, news carries width as well as direction: neutral news raises publication-day volatility. The study also produces a table of measured drift for each event tag, usable as a prior in news-conditioned forecasting models.
\end{abstract}



\maketitle

\section{Introduction}

Every trading day, thousands of news articles are written about
individual stocks, and a growing share of quantitative systems read
them. Two old sayings surround this flow: that news is already priced
in by the time it is published, and that the rumor is bought while the
news is sold. Taken literally, both locate the price move connected to
a news event before and at publication rather than after it.
Watching a live news feed makes the question concrete: often
enough, a stock receives plainly good news and its price falls anyway,
or drifts down for weeks after upbeat coverage. Seeing this repeatedly
was the direct motivation for this study. Whether
that is accurate, for which kinds of news, and by how much are basic
questions about how fast markets absorb public information, and their
answers determine what news can and cannot contribute to forecasting
and trading systems.

These questions have stayed open at the scale of the full news flow
because three layers of measurement infrastructure were missing. The
first is tagged events: an earnings report, a lawsuit, and a
promotional listicle are different economic objects that move prices
differently, so each article must be labeled with the kind of event it
reports, at high accuracy and for millions of articles. The second is
story structure: the same event is syndicated and followed up for days,
so repeated coverage must be grouped into one story before events can
be counted. The third is a benchmark for publicity itself: separating
the effect of a news direction from the effect of merely being in the
news requires knowing what happens to a stock that receives coverage
with no direction at all. The recent literature that applies large
language models to financial news documents genuine return
predictability, but it concentrates on prediction and builds none of
these measurement layers.

This paper builds the three layers and runs the measurement. We
introduce a taxonomy of 17 news tags and five attributes, fixed before this study. Its design drew on a pilot of a subset of large stocks under an earlier system of 63 loosely defined tags, where the behavior of prices around each tag guided which tags were kept, merged, or dropped. A
two-step pipeline tags the full corpus: an Azure-hosted GPT teacher
labels a sample, and a small distilled classifier, refined with active
learning, extends the labels to all 4.57 million articles covering
roughly 3{,}000 US stocks (2023 to 2026) at 95 percent fidelity to the
teacher and negligible cost. A clustering step then groups articles
into stories, so first reports separate from follow-up coverage. On
this foundation we measure, quantitatively: when news is priced in and
to what extent the two sayings hold; what pure publicity does to
prices, as distinct from positive or negative coverage; how news moves
volatility; and how story-driven news differs from fundamental news.
Returns are measured as beta-adjusted abnormal returns around 1.68
million (stock, day, tag) events, with significance from a bootstrap
that resamples trading dates, and with 364{,}405 neutral-sentiment
events serving as the built-in placebo for publicity.

The two expressions hold, and the placebo reveals a third fact that
reinterprets much of the raw picture. The news-aligned move sits
overwhelmingly before and at publication, and for rumor-flagged events,
the directional return is concentrated around the rumor event. A drift that exists with or
without news, the residual of the return benchmark over this sample,
is behind what otherwise looks like widespread post-news reversal;
the placebo design removes it. Net of
it, markets underreact to quantified fundamental news and overreact to
soft stories. Section~\ref{sec:results} develops each finding with the
evidence.

\textbf{Contributions.} Our contribution is measurement and reinterpretation, not methodological novelty. (1) A scalable recipe for labeling
multi-million article corpora by distilling a large language model
teacher into a compact classifier, at 95 percent agreement with its
Azure GPT teacher and negligible cost. (2) The largest tagged,
placebo-controlled measurement of where news-aligned price moves sit in
event time that we are aware of. (3) The background-drift correction,
which reinterprets post-news reversal, the apparent asymmetry between
good and bad news, and a class of news-fading strategies. (4) A
per-tag, per-attribute, per-size drift table usable as a ranking and
importance-sampling prior for news-conditioned forecasting models.

\section{Related work}

Recent work on LLMs and equity news splits into three strands. The first
asks whether LLM-extracted sentiment predicts returns:
\citet{lopezlira2025chatgpt} show ChatGPT headline scores predict
next-day returns, \citet{chen2025deepseek} extend the exercise to market
and macro prediction, \citet{kirtac2024sentiment} trade LLM sentiment
signals, and \citet{iacovides2024finllama} fine-tune open models for the
same task; \citet{kim2024fsa} apply LLMs to fundamental analysis and
\citet{papasotiriou2024ratings} to stock ratings, and
\citet{ghafouri2025risk} probe the behavioral signatures of the models
themselves. \citet{kargarzadeh2024trading} and \citet{ghatak2025alpha}
combine LLM news signals with macroeconomic and technical indicators in
trading frameworks. \citet{chen2024expected} extract news embeddings across 16
markets and find short-horizon news momentum that persists for days in
small stocks, a finding our adjusted earnings and analyst drifts echo
from the event side. \citet{jadhav2025survey} review LLM applications
in equity markets and find the literature concentrated on sentiment
extraction and return prediction; \citet{cao2025survey} draw the same
picture for quantitative investment more broadly, with news entering
models as a predictive feature rather than as an object of
measurement. Our question is
different: we do not build a predictor, we measure where in event time
the news-aligned move sits, and our placebo design shows that a naive
reading of post-news drift would mislead exactly the strategies this
strand builds.

The second strand structures news into events rather than sentiment
scalars. \citet{li2025structured} learn structured event representations
for return prediction, \citet{wang2025stockmem} maintain an event memory
for forecasting, and \citet{liXiang2026janus} train end-to-end
event-driven trading policies. These systems need to know which event
tags matter and how fast their information decays; our measured drift
table is exactly that prior, estimated on 1.68 million events rather
than learned end to end.

The third strand is methodological. \citet{pangakis2024distillation} and
\citet{xia2025candist} show that distilling LLM-generated labels into
small supervised classifiers matches human annotation quality at a
fraction of the cost; our tagging pipeline is an industrial-scale
instance with per-tag validation, and \citet{khaledian2025pcarag} cut
the cost of retrieval over financial text along similar lines. \citet{gao2025lookahead} and
\citet{chen2026nowcasting} document look-ahead risks when LLMs generate
forecasts from text they may have memorized, and
\citet{he2025chronological} train chronologically consistent models to
avoid them; our design sidesteps the problem, since the LLM only assigns
event tags and attributes, never forecasts, and the corpus is produced
daily in live operation.

\section{Data}
\label{sec:data}

\CorpusIntro{} Each day it processes over a quarter million news URLs
concerning the roughly 3{,}000 most-covered US-listed stocks, drawn from
a crawl of more than half a million sources; over the sample this
amounts to more than 70 million raw articles. The 4.57 million
retained here are the important ones: deduplicated, financial in
nature, and directly related to the covered stock universe. For every
crawled article, the title, body, and metadata are read by an LLM that
decides whether the article is genuinely related to the stock and
important enough to keep, assigns a sentiment direction on five levels
from strongly negative to strongly positive, and writes a summary;
retained articles carry these fields along with the publication
timestamp and source domain.\CorpusPrior{} Sentiment and summaries
in the live period are produced day by day, with no access to
subsequent returns. Our event tags were assigned retrospectively, by a
classifier trained later; they are descriptive labels of what an
article reports, not forecasts, and the teacher-only cut in
Section~\ref{sec:map} shows the findings do not depend on the
classifier or its training vintage. The 2023 portion was backfilled in
bulk, is under 3\% of signed events, and shows the same patterns as the
live years.
Prices are daily adjusted closes for 2{,}591 tradeable symbols; 3.4\% of
articles reference symbols without price history (delistings skew
small), and 9.7\% of events are skipped for missing price windows.

\section{Method}
\label{sec:method}

Our method has three stages: tag every article with a distilled LLM
tagger, group articles into stories and events, and measure abnormal
returns around those events against a placebo.

\subsection{Distilled LLM tagging}

A frozen prompt presents 17 event-tag definitions, five binary
attribute definitions (scheduled, forward looking, primary source,
quantified, rumor), and four example articles with their completed
labels to a small GPT teacher
(gpt-5-mini, hosted on Azure) with a strict JSON schema, at about
\$0.0002 per article. The taxonomy covers real corporate events
(earnings, guidance, analyst actions, launches, M\&A, legal and
regulatory, and so on) and deliberately includes two junk categories,
price commentary and promotional content, so that articles describing
price action or promotion have somewhere to go other than a real event
tag. The teacher labeled 29{,}472 random articles; a distilroberta-base
student (82M parameters, seven heads: primary tag, secondary tag, five
attributes) trained on these labeled 600{,}000 fresh articles; the
100{,}000 lowest-confidence articles went back to the teacher
(uncertainty sampling, with a quota for rare tags); and the student was
retrained on the merged 129{,}463 labels (29{,}472 + 99{,}991; nine of the 100{,}000 produced no usable label). On a held-out 10{,}000-article
sample labeled independently by both, the student agrees with the
teacher on 87.5\% of primary tags overall and 94.0\% at confidence
$\geq0.8$ (85\% of articles); attribute agreement is 92--99\%.
Deployment uses the classifier everywhere, routes the 15\%
low-confidence tail back to the teacher, and keeps teacher labels where
they exist, for an estimated 95\% corpus-wide teacher fidelity at a cost of \$29.34 for the 139{,}464 teacher calls that trained and validated the classifier, and \$156.29 in total including the routed tail. Every article records its label
source, which powers a robustness check below. Inference runs at
125--160 articles per second on a MacBook Pro laptop (M4 Pro, 24\,GB
RAM).

\subsection{From articles to events}

Bundling repeated coverage into distinct events is a hard problem in
its own right, and we solve it with a two-step, embedding-based
clustering. Within each stock-day, articles are clustered on the
cosine similarity of their title-and-summary embeddings (threshold
0.80), restricted to the same primary tag; each day-cluster is then
matched against the stock's running stories, so an event continues
across days when later coverage stays close to the story's frozen
centroid. The clustering quantifies how much news repeats itself: it
reveals that 55\% of all articles are follow-up coverage of a story
already running, and it flags each article as the first report (NEW)
or follow-up. How long a story may stay open depends on its tag: we
calibrated per-tag lifetimes by tracking how the similarity of later
same-story coverage to the first-day centroid decays with story age,
and capped each tag where continued matches become rare. M\&A and
legal sagas stay open for up to 90 days, leadership and operations
stories for weeks, commentary for two days; measured persistence per
tag is reported in Table~\ref{tab:persist}. For the event study, an
event is a (stock, trading day, tag) aggregate: articles published
after 16:00 New York time roll to the next trading day, and the
article count is retained as an attention measure. The event's impact
comes from the vendor sentiment: each article carries one of five
levels, and the event's impact averages its articles. An event is
neutral when every one of its articles is Neutral, and signed
otherwise. The drift tables use only the sign of the impact. This
yields 1{,}862{,}297 events, of which 1{,}681{,}657 are scored against
prices: 1{,}317{,}252 signed (73\% positive) and 364{,}405 neutral.

\begin{table}[t]
\caption{Story persistence by tag, measured from the clustering:
number of stories and mean lifetime in trading days.}
\label{tab:persist}
\centering\small
\resizebox{\columnwidth}{!}{\begin{tabular}{lrr|lrr}
\toprule
Tag & Stories & Days & Tag & Stories & Days \\
\midrule
M\&A deal & 22,596 & 12.9 & Macro through stock & 322,666 & 1.1 \\
Legal, regulatory & 46,846 & 10.6 & Financing, dilution & 11,497 & 0.9 \\
Insider, ownership & 119,394 & 4.0 & Product launch & 159,672 & 0.9 \\
Analyst action & 99,847 & 2.5 & Competition & 32,131 & 0.8 \\
Leadership & 53,794 & 2.4 & Partnership & 103,199 & 0.8 \\
Operations, supply & 89,071 & 2.3 & Price commentary & 220,451 & 0.6 \\
Earnings results & 89,433 & 2.3 & Other corporate & 100,891 & 0.5 \\
Capital returns & 21,796 & 1.6 & Promotional & 93,974 & 0.4 \\
Guidance, outlook & 24,343 & 1.4 & & & \\
\bottomrule
\end{tabular}
}
\end{table}

\subsection{Measurement design}
\label{sec:design}

Let $r_t$ be the stock's daily return and $r_{m,t}$ the daily return
of the S\&P 500 index, measured through SPY, the exchange-traded fund
that tracks it. The abnormal return is $AR_t = r_t - \beta_t\,
r_{m,t}$ with $\beta_t$ a rolling 252-day OLS beta of the stock
against the S\&P 500, as of day $t$ (minimum 126 observations, missing
betas set to 1). Around each event day we cumulate
$AR$ over four windows: days $-5..0$ (which includes day 0), day 0, days $+1..+5$, and days
$+6..+20$. Signed events are stacked by sentiment sign
$s\in\{-1,+1\}$, so a move in the direction of the news counts as
positive whether the news was good or bad. Significance comes from a
cluster bootstrap that resamples trading dates (5{,}000 draws), because
events on the same day are cross-sectionally correlated. For size
buckets, stocks are split each year into three equal-sized groups
(small, mid, large) by dollar volume.

\textbf{The placebo and the adjusted estimator.} A single-beta
benchmark is deliberately simple: transparent, replicable, and
standard. No fixed benchmark tracks every stock perfectly, and
whatever residual drift a benchmark leaves behind flows into every
event window measured against it. The design therefore includes a
placebo rather than a bet on the benchmark. Neutral events carry
coverage without direction, so their drift measures what happens to
comparable stocks of the same size around news days in general. Call
that the baseline $b_{k,w}$, the mean abnormal
return of neutral events in size bucket $k$ over window $w$. When we
want the effect of the news direction itself, we subtract the baseline
from the event's abnormal return before applying the sign:
$s\,(AR_w - b_{k,w})$. Throughout the paper, \emph{adjusted} means
exactly this, the move in the news direction in excess of what
comparable stocks drifted anyway, and \emph{raw} means no
subtraction. Because the baseline is measured against the same
benchmark as the events, any benchmark misfit appears on both sides
of the subtraction and cancels: a richer factor model would shrink
the baseline itself but leave the adjusted estimates essentially
unchanged. Raw and adjusted columns are reported side by side.

Two implementation details matter for inference. First, the baseline
is pooled across event tags within a size bucket, which assumes
coverage drift is tag-independent. A tag-matched variant, which
subtracts each tag's own neutral baseline per bucket, leaves the
earnings continuation and the macro, leadership, and competition
reversals essentially unchanged, while the capital-returns estimate
shrinks and the launch estimate crosses zero; neutral coverage of
those two tags selects unusual stocks (dividend payers, promoted
small caps), so their magnitudes are more model-dependent than their
signs. Second, the baseline is estimated once, on the full neutral
sample; a bootstrap that re-estimates it inside every draw gives the
same or slightly stronger significance for every headline tag, so
ignoring baseline noise in the reported p-values is conservative.

\textbf{What this design can and cannot claim.} Sentiment is assigned
to articles, and articles are written about moves as well as before
them, so the pre-publication drift mixes genuine anticipation with
reporting on moves that already happened. We quantify this with the
NEW-only rerun: restricting to first reports shrinks the earnings
pre-window from $+1.41\%$ to $+1.02\%$, so follow-up coverage inflates
measured anticipation by roughly a third, and all post-publication
conclusions are unchanged. Articles can also describe moves that already happened, so sentiment labels may partly encode realized returns. Our study is a descriptive account of where the news-aligned move sits in event time, not a causal claim about news moving prices.

\section{Results}
\label{sec:results}

Our main findings are as follows. First, the news-aligned move sits
before and at publication, and rumor-flagged events show the buy the
rumor, sell the news pattern literally (Section~\ref{sec:priced}).
Second, a background drift that exists with or without news
explains most of the raw post-news picture
(Section~\ref{sec:baseline}). Third, net of the baseline, quantified
fundamental news drifts and soft news reverses, and the pattern survives
every robustness cut (Section~\ref{sec:map}). The remaining sections
measure source differences, volatility, and economic significance.

\subsection{The move sits before and at publication}
\label{sec:priced}

\begin{figure*}[t]
\includegraphics[width=\textwidth]{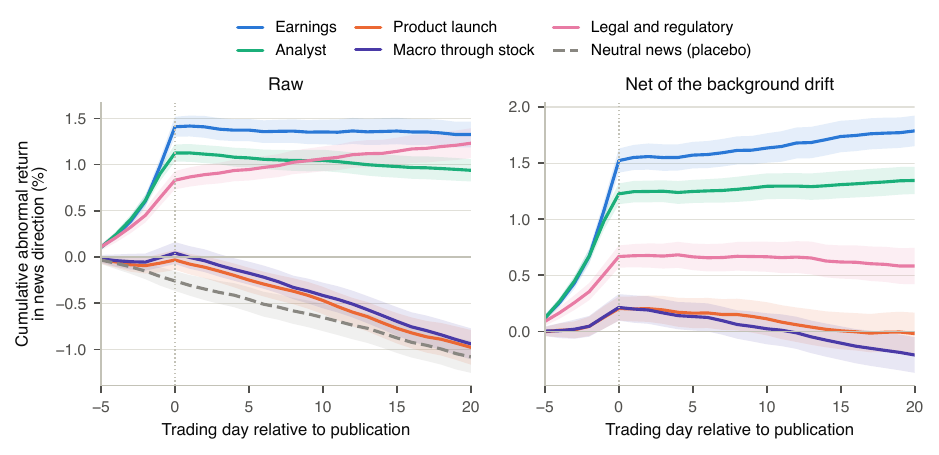}
\caption{Cumulative abnormal return in the news direction around
publication (day 0), five event tags plus the neutral placebo. Left:
raw. Right: net of the background drift. Bands are $\pm2$
date-clustered standard errors. Raw curves show the priced-in signature
(the move precedes publication) and near-universal post-news decay
tracking the placebo. Adjusted curves separate genuine continuation
(earnings, analyst) from genuine reversal (launches, macro), and flatten
legal and regulatory news to the baseline.}
\label{fig:curves}
\end{figure*}

Table~\ref{tab:main} is the master table; Figure~\ref{fig:curves} (left)
shows the event-time picture. Almost every pre and day-0 column in the raw
table is positive: whatever the market does around news, it agrees in
direction with the news before the news is out. For the quantified
fundamental tags the pattern is extreme: earnings $+1.41\%$ pre and
$+0.42\%$ on the day against $-0.05\%$ afterwards; analyst actions
$+1.12\%$ and $+0.22\%$ against $-0.13\%$; price commentary, guidance,
and competition all show the same shape. A useful summary comes
from the running total plotted in Figure~\ref{fig:curves}: starting
five days before publication, add up each day's average abnormal
return in the news direction. Pooled across all signed events, this
running total reaches $+0.579\%$ by the close of publication day, then falls back and ends day $+20$ at $+0.204\%$. The ratio of the two levels is 2.84: by the closing bell of publication day the market had
already moved almost three times as far as where it would stand a
month later, and the extra ground was given back over the following
weeks. The ratio measures how complete the move was, not how large:
earnings events have the largest moves but almost no give-back, so
their ratio is 1.06, against 1.20 for analyst actions and 1.08 for
guidance; at or above one throughout, the move was finished, or more
than finished, when publication day ended. The pooled ratio is higher
than any single tag's because the pooled after-window also inherits
the background drift examined in Section~\ref{sec:baseline}.

\begin{table*}[t]
\caption{Signed abnormal drift by event tag, sorted by adjusted
post-event drift. All columns are percent, in the direction of the news.
``\%pos'' is the share of positive-sentiment events. ``adj.'' subtracts
the baseline of the event's size bucket before signing
(Section~\ref{sec:design}); $p_{\text{adj}}$ is the date-bootstrap
p-value of the adjusted days $+6..+20$ drift.}
\label{tab:main}
\centering\small
\begin{tabular}{lrrrrrrrr}
\toprule
Event tag & $n$ & \%pos & $-5..0$ & day 0 & $+1..+5$ & $+6..+20$ & adj.\ $+6..+20$ & $p_{\text{adj}}$ \\
\midrule
Capital returns & 26,273 & 90 & +0.22 & +0.06 & -0.01 & -0.14 & +0.35 & $<$0.001 \\
Earnings results & 107,174 & 72 & +1.41 & +0.42 & -0.04 & -0.05 & +0.22 & $<$0.001 \\
Guidance and outlook & 24,828 & 69 & +1.43 & +0.56 & -0.02 & -0.09 & +0.13 & 0.046 \\
Analyst action & 134,464 & 70 & +1.12 & +0.22 & -0.05 & -0.13 & +0.10 & 0.012 \\
Price commentary & 174,087 & 61 & +1.55 & +0.43 & -0.02 & -0.03 & +0.09 & 0.053 \\
Promotional content & 55,230 & 96 & -0.01 & 0.00 & -0.11 & -0.43 & +0.08 & 0.274 \\
Financing and dilution & 9,161 & 78 & +0.77 & +0.30 & +0.02 & -0.28 & +0.07 & 0.554 \\
Operations and supply & 59,613 & 57 & +0.45 & +0.14 & +0.02 & -0.05 & +0.04 & 0.314 \\
Insider and ownership & 238,236 & 70 & +0.16 & +0.04 & -0.06 & -0.21 & +0.03 & 0.459 \\
M\&A deal & 25,962 & 84 & +0.17 & +0.04 & -0.23 & -0.46 & -0.05 & 0.498 \\
Partnership, customer & 76,848 & 97 & +0.08 & +0.05 & -0.19 & -0.63 & -0.08 & 0.317 \\
Legal and regulatory & 46,057 & 17 & +0.83 & +0.19 & +0.11 & +0.29 & -0.08 & 0.220 \\
Other corporate & 47,485 & 82 & 0.00 & +0.04 & -0.16 & -0.48 & -0.09 & 0.200 \\
Competition & 20,744 & 40 & +0.90 & +0.32 & -0.09 & -0.08 & -0.17 & 0.026 \\
Leadership, governance & 30,388 & 75 & +0.28 & +0.08 & -0.10 & -0.51 & -0.18 & 0.016 \\
Product launch & 101,425 & 96 & -0.03 & +0.03 & -0.22 & -0.73 & -0.18 & 0.022 \\
Macro through stock & 139,277 & 82 & +0.04 & +0.05 & -0.22 & -0.76 & -0.34 & $<$0.001 \\
\bottomrule
\end{tabular}

\end{table*}

The rumor attribute makes the expression measurable
(Table~\ref{tab:rumor}). Among 18{,}618 rumor-flagged signed events,
94\% are followed by a non-rumor event with the same tag within 60
trading days (median gap two trading days). The rumor day delivers $+0.36\%$ in
the rumor's direction. Between rumor and confirmation the stock adds
nothing ($-0.09\%$). The confirmation day itself is worth $+0.01\%$,
and days +6 to +20 after confirmation give back $-0.06\%$. M\&A rumors
are the sharpest case: $+0.24\%$ on rumor day, then $-0.32\%$ into the
confirmation, $-0.10\%$ on the news, and $-0.37\%$ after it. Note that the linked event is the next same-tag, non-rumor event for the same stock; therefore these numbers represent returns around linked pairs. Consistent with this, events preceded by a
same-tag rumor in the prior month show no additional drift after the
news.

\begin{table}[t]
\caption{Buy the rumor, sell the news. Mean abnormal return in the
rumor's direction (percent) at each stage, for rumor-flagged events with
a same-tag non-rumor event within 60 trading days.}
\label{tab:rumor}
\centering\small
\resizebox{\columnwidth}{!}{\begin{tabular}{lrrrrrr}
\toprule
Sample & $n$ & Rumor day & Rumor$\to$news & News day & $+1..+5$ & $+6..+20$ \\
\midrule
All rumor events & 17,510 & +0.36 & -0.09 & +0.01 & -0.14 & -0.06 \\
M\&A deal & 3,335 & +0.24 & -0.32 & -0.10 & -0.16 & -0.37 \\
Legal and regulatory & 1,452 & +0.55 & -0.01 & -0.01 & +0.17 & -0.17 \\
Product launch & 2,316 & +0.13 & +0.12 & -0.08 & -0.35 & +0.02 \\
\bottomrule
\end{tabular}
}
\end{table}

\subsection{The background drift and the placebo}
\label{sec:baseline}

\begin{figure}[t]
\includegraphics[width=0.92\columnwidth]{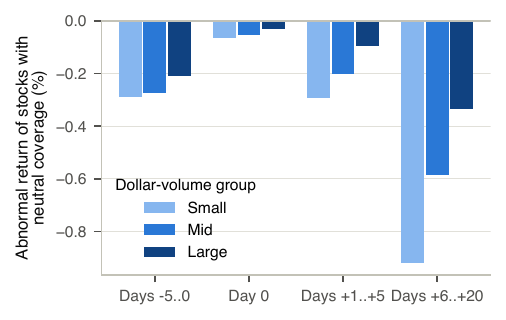}
\caption{The background drift: abnormal returns of stocks receiving
neutral news, by window and size group. Quiet stock-days with no
recent news drift similarly (Section~\ref{sec:baseline}): the drift
is the residual of the single-beta benchmark over this sample, not an
effect of publicity, and the placebo adjustment removes it either
way.}
\label{fig:baseline}
\end{figure}

The placebo was designed as a calibration check and became a
finding. Neutral events measure what happens to a stock around news
days that carry no direction; quiet stock-days, defined as days with
no event for the stock and none in the prior five trading days,
measure the same windows with no news at all. Both drift down. Over
the following month, stocks with neutral coverage trail their beta
benchmark by $-0.92\%$ among small caps, $-0.58\%$ among mid caps,
and $-0.34\%$ among large caps; quiet days drift $-0.74\%$,
$-0.88\%$, and $-0.59\%$ in the same buckets
(Figure~\ref{fig:baseline}). The two sets of numbers tell one story.
The drift is not caused by publicity: it is the residual of the
single-beta benchmark over this sample, present with news and
without, of the same order across size groups, with covered small
caps adding a modest $0.2\%$ per month on top and covered mid and
large caps adding none. This is exactly why the design subtracts a
placebo instead of trusting any benchmark: news studies inherit a
drift that exists with or without news, and any measurement or
strategy that scores news direction without a placebo of comparable
stocks will misread that drift as directional information.

To see how this distorts directional measurements, take an average
positive-news event in a small cap. Its raw post-event drift is the sum
of two parts: the market's reaction to the news direction, plus the
$-0.9\%$ that any covered small cap drifts over the following month
anyway. Measured in the direction of the news, that second part shows
up as \emph{reversal} of good news. Now take a negative-news event in
the same stock: the same $-0.9\%$ drift now points in the same
direction as the news, so it shows up as \emph{continuation} of bad
news. One underlying drift, two apparent behaviors. Because 73\% of
signed events are positive, the pooled average is dominated by the
first case, which is why the raw table looks like widespread reversal.

\begin{figure}[t]
\includegraphics[width=0.92\columnwidth]{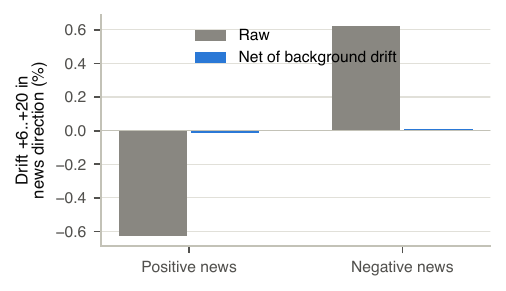}
\caption{The asymmetry illusion. Raw post-news drift suggests good news
reverses and bad news continues. Once the baseline is
subtracted, both collapse to zero: the apparent asymmetry is the
baseline interacting with the 73\% positive share of news sentiment.}
\label{fig:asym}
\end{figure}

The adjusted estimator of Section~\ref{sec:design} removes exactly this
confound, and three apparent anomalies disappear at once
(Figure~\ref{fig:asym}). The apparent asymmetry between good and bad
news: raw drift over days +6 to +20 is $-0.62\%$ after positive news
(looks like overreaction to good news) and $+0.63\%$ after negative
news (looks like underreaction to bad news); after the baseline is
subtracted, both are within $0.02\%$ of zero ($p\approx0.9$). The size gradient:
small caps appear to reverse 2.4 times more than large caps in the raw
data ($-0.41\%$ against $-0.17\%$ per month), and adjusted, the
gradient is gone. And
the one apparent
continuation winner in the raw table, legal and regulatory news
($+0.29\%$, $p<0.001$): legal news is one of only two tags where most events are negative (83\%; competition is the other, at 60\%), so for this tag the baseline reads as
continuation rather than reversal; adjusted, legal news drifts
$-0.08\%$ ($p=0.22$), indistinguishable from any other covered stock.

\subsection{What survives: the tag map}
\label{sec:map}

\begin{figure}[t]
\includegraphics[width=\columnwidth]{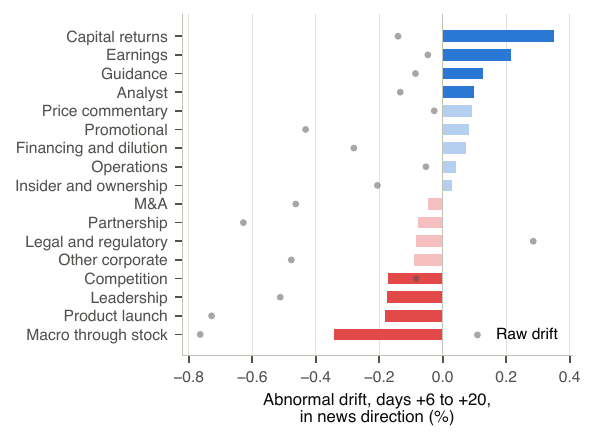}
\caption{The tag map: adjusted drift over days +6 to +20 in the news
direction (bars; pale bars not significant at 5\%) against raw drift
(circles). Adjustment moves every tag toward zero and reorders them:
quantified fundamentals drift, soft news reverses.}
\label{fig:map}
\end{figure}

Figure~\ref{fig:map} and the adjusted columns of Table~\ref{tab:main}
give the residual structure, and it is orderly. On the continuation
side sit the quantified fundamental tags: capital returns $+0.35\%$,
earnings $+0.22\%$, guidance $+0.13\%$, analyst actions $+0.10\%$ over
days +6 to +20, all significant; the earnings figure is the familiar
post-earnings-announcement drift recovered at corpus scale from tagged
news alone. On the reversal side sit the soft, attention-driven tags:
macro read-throughs $-0.34\%$, product launches $-0.18\%$, leadership
stories $-0.18\%$, competition $-0.17\%$. Launch and partnership
coverage is 96--97\% positive, the most promotional corner of the
corpus, and it is precisely where prices overshoot.

Table~\ref{tab:robust} stress-tests the map with three cuts. The first
keeps only NEW articles (1.09M events), removing all follow-up
coverage. The second addresses simultaneous stories: a covered
stock-day typically carries more than one event tag at once (the
median is two), so drift attributed to a launch could in principle
belong to a co-occurring earnings story; restricting to stock-days with
exactly one event tag removes that contamination. The third addresses
classifier error: recomputing only on events whose articles were
labeled directly by the GPT teacher, bypassing the distilled classifier
entirely, tests whether tagging mistakes create the patterns. Across
the three cuts the headline tags keep their signs and similar
magnitudes, with one exception: analyst actions flip to $-0.05\%$ on
single-tag days. The exception has a natural reading: analyst notes
rarely appear without other same-day coverage (fewer than seven
percent of analyst events sit on single-tag days), so that cell is a
small and unusual subsample. Single-tag days otherwise strengthen the
survivors (capital returns $+0.47\%$, launches $-0.48\%$, macro
$-0.52\%$). By year, the raw pooled reversal halves from 2024
($-0.49\%$) to 2026 ($-0.18\%$), and the adjusted pooled drift is small in every year ($-0.16\%$ to $+0.07\%$). Finally, because
seventeen tags are tested at once, we apply Benjamini-Hochberg
control to the adjusted p-values of Table~\ref{tab:main}: at a 5
percent false discovery rate the macro reversal and the
capital-returns and earnings continuations survive, the analyst, leadership, launch, and competition results sit at the boundary ($q$ between 0.05
and 0.07), and the guidance continuation is suggestive rather than
decisive.

\begin{table}[t]
\caption{Robustness of the adjusted days $+6..+20$ drift (percent).
Columns: all events; NEW articles only; stock-days with one event tag;
events labeled directly by the GPT teacher.}
\label{tab:robust}
\centering\small
\resizebox{\columnwidth}{!}{\begin{tabular}{lrrrr}
\toprule
Event tag & All events & NEW only & Single-tag days & Teacher labels \\
\midrule
Capital returns & +0.35 & +0.33 & +0.47 & +0.34 \\
Earnings results & +0.22 & +0.11 & +0.17 & +0.11 \\
Guidance and outlook & +0.13 & +0.06 & +0.34 & +0.08 \\
Analyst action & +0.10 & +0.07 & -0.05 & +0.08 \\
Legal and regulatory & -0.08 & -0.07 & -0.25 & -0.01 \\
Product launch & -0.18 & -0.18 & -0.48 & -0.14 \\
Leadership, governance & -0.18 & -0.14 & -0.26 & -0.17 \\
Macro through stock & -0.34 & -0.26 & -0.52 & -0.24 \\
\bottomrule
\end{tabular}
}
\end{table}

Attributes slice the same way (not tabulated for space): scheduled
events show more anticipation and slightly positive adjusted drift
($+0.08\%$, $p=0.03$), unscheduled events none; quantified articles show no drift, unquantified ones reverse ($-0.12\%$, $p=0.01$); rumor-flagged
events show the largest pre-windows. Publication timing matters for the
day itself: articles published during market hours land on a day that
moves 1.18 times the stock's normal absolute move, against 1.05 for
overnight publications.

\subsection{Sources are not interchangeable}

Grouping events by the kind of outlet that wrote them
(Table~\ref{tab:sources}) separates anticipation from information.
Retail-analysis sites, portals, mainstream media, and aggregators all
show $+0.8\%$ to $+1.1\%$ of pre-publication drift: they write about
moves in progress. Press wires, which carry company and regulator
releases, show none at all ($-0.06\%$): the release is the event, not
commentary on one. Wire-sourced positive tilt also reverses hardest
($-0.19\%$ adjusted, marginal), consistent with promotional press
releases overselling. At the level of individual outlets, which we
leave unnamed, the two with the most promotional catalogs show
economically large negative adjusted drift ($-0.4\%$ to $-0.8\%$ per
month), suggesting a source-quality signal that survives the baseline;
we leave a full source-reputation study to future work.

\begin{table}[t]
\caption{News source groups. NEW share is the fraction of articles that
are the first report of their story; day-0 $|AR|$ ratio is the
publication-day absolute move as a multiple of the stock's normal day.
Events here are (stock, day, tag, source group) aggregates, so a story
covered by several source groups appears once per group; counts
therefore exceed the number of signed events.}
\label{tab:sources}
\centering\small
\resizebox{\columnwidth}{!}{\begin{tabular}{lrrrrrr}
\toprule
Source group & $n$ & NEW share & $-5..0$ & day 0 & day-0 $|AR|$ ratio & adj.\ $+6..+20$ \\
\midrule
Retail analysis & 224,122 & 50\% & +1.07 & +0.41 & 1.28 & +0.09 \\
Portals & 294,359 & 55\% & +0.92 & +0.28 & 1.22 & +0.11 \\
Mainstream media & 70,616 & 49\% & +0.86 & +0.38 & 1.26 & +0.01 \\
Aggregators & 544,809 & 38\% & +0.80 & +0.21 & 1.12 & +0.06 \\
Other & 475,705 & 63\% & +0.46 & +0.17 & 1.07 & -0.12 \\
Press wires & 117,517 & 60\% & -0.06 & +0.05 & 1.17 & -0.19 \\
\bottomrule
\end{tabular}
}
\end{table}

\subsection{Second moments: news carries width}

Neutral news is directionally empty but not informationless. Neutral
guidance events come with publication-day moves 1.29 times the stock's
normal day and 1.23 times over the following week; neutral earnings
1.11 times. Attention scales the effect: days with ten or more articles
move 1.36 times normal against 1.05 for single-article days. All
of these are ratios of a stock's absolute move to its own typical
day, so the background drift of Section~\ref{sec:baseline}, tiny at
the daily scale, does not enter them. A distributional forecaster
should therefore read event tags as width signals, widening on scheduled disclosure events, independent of
direction.

\subsection{Economic significance}

\begin{figure}[t]
\includegraphics[width=\columnwidth]{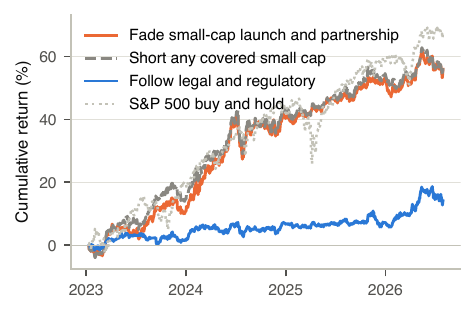}
\caption{Cumulative returns of the calendar-time portfolios,
each hedged with a beta-sized index position, next to an S\&P 500 buy
and hold. The sentiment-fading strategy (orange) is indistinguishable
from shorting every covered small cap (dashed): the news direction
adds nothing.}
\label{fig:port}
\end{figure}

Table~\ref{tab:port} is an in-sample diagnostic of the two raw patterns a practitioner would be tempted to trade, not evidence of a tradable strategy. Both are calendar-time portfolios built the same
way: a position opens at the close of day +5 after a qualifying event
and closes at the close of day +20 (the window where the drift lives),
all open positions are equally weighted and rebalanced daily, and
returns are measured on beta-hedged abnormal returns, so the portfolios
are market-neutral overlays. Concretely, each position is a
stock leg plus an offsetting beta-sized position in the index, so the
returns and Sharpe ratios reported here are the realized returns of
that hedged implementation. The first strategy \emph{fades} small-cap
launch and partnership news: after a positive-sentiment launch or
partnership event in a small cap it shorts the stock, after a
negative-sentiment one it buys, betting that the sentiment move
reverses. It earns 15.8\% annualized gross with a Sharpe ratio of 1.35
and survives 20 bps of cost per side (Sharpe 0.77). The benchmark
strategy ignores sentiment entirely: it shorts \emph{every} small-cap
stock that appeared in the news at all, neutral or signed, over the
same windows. It earns 15.9\% with the same Sharpe of 1.35, and the two
cumulative curves lie on top of each other (Figure~\ref{fig:port}). The
fading strategy's entire edge is the background drift; the news
direction contributes nothing. Following legal and regulatory news
direction, the raw table's continuation candidate, earns a Sharpe of
0.66 gross that dies at 10 bps, as the composition argument predicts.
Both headline strategies are predominantly short books, and the
cost model charges only a flat fee per side: borrow fees and locate
availability are not modeled, and for small caps they would consume a
further slice of the returns. The residual tag-map alphas (earnings
continuation at $+0.22\%$ per
event over 15 days) are real but thin relative to plausible costs. 

\begin{table}[t]
\caption{Calendar-time portfolios, days +6 to +20, abnormal returns,
2023 to 2026. Net Sharpe ratios charge the stated cost per side.}
\label{tab:port}
\centering\small
\resizebox{\columnwidth}{!}{\begin{tabular}{lrrrrrr}
\toprule
Strategy & Events & Ann.\ ret.\ & Sharpe & Sharpe 5\,bps & 10\,bps & 20\,bps \\
\midrule
Fade small-cap launch, partnership & 52,149 & 15.8\% & 1.35 & 1.20 & 1.06 & 0.77 \\
Short any covered small cap & 260,472 & 15.9\% & 1.35 & 1.21 & 1.07 & 0.78 \\
Follow legal and regulatory & 46,057 & 4.1\% & 0.66 & 0.39 & 0.12 & -0.43 \\
\bottomrule
\end{tabular}
}
\end{table}

\section{Implications for news-conditioned forecasting}

These measurements bear directly on the design of news-conditioned
forecasting systems, in three ways. First, ranking: when a day's news
must be compressed into a short context, event lines should be ordered
by the measured per-tag, per-attribute, per-size drift priors rather
than by source reputation heuristics. Second, representation: since
direction is mostly priced in at publication while width effects
persist, the text should be encoded as tagged, flagged event lines
(tag, scheduled, rumor, primary source, NEW or follow-up age) rather
than as sentiment scalars; the tag and flags carry the durable
information. Third, evaluation: any claimed news alpha should be
benchmarked against a coverage-presence baseline of the kind measured
here, or it will rediscover the baseline and call it
signal.

\section{Limitations}

Five limitations frame how far these results reach, and each points at
follow-up work. The first concerns labels. Sentiment and summaries come
from the corpus vendor's LLM, and our event tags are layered on top by
a distilled classifier that disagrees with its teacher on about one
article in eight. Those disagreements concentrate in the two junk
categories, price commentary and promotional content, where a wrong
label matters least, and the teacher-only check in
Table~\ref{tab:robust} shows the conclusions do not depend on them.
Still, a mislabeled article can end up attached to the wrong story, so
cleaner labels would sharpen the clustering as well. The second
concerns the return benchmark. Each stock is hedged with a single
market beta, so style effects such as size are not explicitly modeled;
the placebo adjustment absorbs much of this, because the baseline is
measured within size buckets, but a formal multi-factor treatment is a
natural next step. Third, timing. Daily closes cannot separate the
trading-hours reaction from the overnight gap, and the sample covers
2023 to 2026, a single market regime whose year-by-year decay suggests
the residual patterns are already fading; the 2023 slice is backfilled
rather than live. Fourth, the corpus is public news, the news
available on the open internet, which is what most market
participants actually read. It is in the nature of public information
that by the time an event is written up in an article anyone can
access, some of it has already moved through filings, terminals, and
professional channels; this holds for any collection of public news,
and part of what we measure as anticipation is this property of
public information itself. Public news is largely priced by the time
it appears, but it is not empty: \citet{kargarzadeh2024trading}
builds trading strategies on public news of this kind, combined with
macroeconomic indicators in a momentum-style framework, and reports
good results, and our tag map shows where such residual information
lives. Finally, the economic significance exercise is
deliberately simple: fixed windows, equal weights, flat costs. It is
built to price the measured patterns, not to be a trading system.
Richer strategies can certainly be developed on this data; here the
focus is on how news is priced, and we leave strategy design to future
work.

\section{Conclusion}

We asked when news is priced in, and we answered it at the
scale of the full news flow. Both market sayings describe the data well.
Across 1.68 million tagged events, the price move connected to a
news event lives before and at publication: anticipation builds over
the preceding days, publication day absorbs what remains, and the
weeks after add nothing that was not already there. For rumor-flagged events, the directional return is concentrated around the rumor, with little added at the next linked event.

The measurement also shows how easily news gets more credit
than it deserves. Stocks drifted below a standard beta benchmark
across this sample whether they were in the news or not, and without
a placebo of comparable stocks that drift reads as reversal of good
news, continuation of bad news, and profitable-looking strategies
that fade news sentiment. Subtracting the placebo makes these
illusions disappear together, and what remains is our
simplest lesson about how news is priced: markets underreact to
numbers and overreact to stories. Hard, quantified disclosures keep
drifting in their own direction for weeks after publication, and soft narrative coverage gives back its move.

For anyone building forecasting or trading systems on news, the
durable information is the kind of event, the fact and intensity of
coverage, and its width.

\begin{acks}
This work was carried out at TailState Intelligence Ltd. The
NewsWitch news corpus and its data infrastructure were provided by
Zanista AI Ltd. We thank the Zanista AI team for data access and
support.
\end{acks}

\bibliographystyle{ACM-Reference-Format}
\bibliography{refs}

\end{document}